\documentclass[conference]{IEEEtran}
\IEEEoverridecommandlockouts
\usepackage{cite}
\usepackage{amsmath,amssymb,amsfonts}
\usepackage{algorithm,algorithmic}
\usepackage[caption=false]{subfig}
\usepackage{graphicx}
\usepackage{textcomp}
\usepackage{xcolor}
\usepackage{pifont}
\usepackage{hyperref}
\usepackage{microtype}
\usepackage{booktabs} 
\usepackage{ifthen}
\usepackage{fancyhdr}
\usepackage{multirow}
\usepackage{etoolbox}
\usepackage{booktabs}
\usepackage{verbatim}
\usepackage{tikz}
\usetikzlibrary{arrows, arrows.meta}
\usetikzlibrary{external}
\def\BibTeX{{\rm B\kern-.05em{\sc i\kern-.025em b}\kern-.08em
    T\kern-.1667em\lower.7ex\hbox{E}\kern-.125emX}}

\newcommand{\final}{1}

\definecolor{johnColor}{rgb}{0,0.5,0.7}
\newcommand{\JDO}[1]{{\color{johnColor} JDO\@: #1}}
\ifthenelse{\equal{\final}{1}}
{
  \renewcommand{\JDO}[1]{}
}
{}

\definecolor{zhongyiColor}{rgb}{0.5,0.7,0}
\newcommand{\zhongyi}[1]{{\color{zhongyiColor} zhongyi\@: #1}}
\ifthenelse{\equal{\final}{1}}
{
  \renewcommand{\zhongyi}[1]{}
}
{}

\newcommand{\warpsc}{\textsc{warp}}

\newcommand{\op}[1]{\emph{#1}}

\renewcommand{\headrulewidth}{0pt}
\makeatletter
\patchcmd{\@makecaption}
  {\scshape}
  {}
  {}
  {}
\makeatletter
\patchcmd{\@makecaption}
  {\\}
  {.\ }
  {}
  {}
\makeatother

\newcommand{\pluseq}{\mathrel{{+}{=}}}

\newcommand{\cmark}{\ding{51}}%
\newcommand{\xmark}{\ding{55}}%

\begin{document}

\title{Building a Performance Model for Deep Learning Recommendation Model Training on GPUs
}

\author{\IEEEauthorblockN{Zhongyi Lin}
\IEEEauthorblockA{\textit{Department of Elect.\ \& Comp.\ Engr.} \\
\textit{University of California, Davis}\\
Davis, California \\
zhylin@ucdavis.edu}
\and
\IEEEauthorblockN{Louis Feng}
\IEEEauthorblockA{\textit{Meta Platforms, Inc.}\\
Menlo Park, California \\
lofe@meta.com}
\and
\IEEEauthorblockN{Ehsan K. Ardestani}
\IEEEauthorblockA{\textit{Meta Platforms, Inc.}\\
Menlo Park, California \\
ehsanardestani@meta.com}
\and
\IEEEauthorblockN{Jaewon Lee}
\IEEEauthorblockA{\textit{Meta Platforms, Inc.}\\
Menlo Park, California \\
jaewon@meta.com}
\and
\IEEEauthorblockN{John Lundell}
\IEEEauthorblockA{\textit{Meta Platforms, Inc.}\\
Menlo Park, California \\
jlundell@meta.com}
\and
\IEEEauthorblockN{Changkyu Kim}
\IEEEauthorblockA{\textit{Meta Platforms, Inc.}\\
Menlo Park, California \\
ckkim@meta.com}
\and
\IEEEauthorblockN{Arun Kejariwal}
\IEEEauthorblockA{\textit{Meta Platforms, Inc.}\\
Menlo Park, California \\
akejariwal@meta.com}
\and
\IEEEauthorblockN{John D. Owens}
\IEEEauthorblockA{\textit{Department of Elect.\ \& Comp.\ Engr.} \\
\textit{University of California, Davis}\\
Davis, California \\
jowens@ucdavis.edu}
}

\maketitle

\begin{abstract}
We devise a performance model for GPU training of Deep Learning Recommendation Models (DLRM), whose GPU utilization is low compared to other well-optimized CV and NLP models. We show that both the device active time (the sum of kernel runtimes) but also the device idle time are important components of the overall device time. We therefore tackle them separately by (1)~flexibly adopting heuristic-based and ML-based kernel performance models for operators that dominate the device active time, and (2)~categorizing operator overheads into five types to determine quantitatively their contribution to the device active time. Combining these two parts, we propose a critical-path-based algorithm to predict the per-batch training time of DLRM by traversing its execution graph. We achieve less than 10\% geometric mean average error (GMAE) in all kernel performance modeling, and 4.61\% and 7.96\% geomean errors for GPU active time and overall E2E per-batch training time prediction with overheads from individual workloads, respectively. A slight increase of 2.19\% incurred in E2E prediction error with shared overheads across workloads suggests the feasibility of using shared overheads in large-scale prediction. We show that our general performance model not only achieves low prediction error on DLRM, which has highly customized configurations and is dominated by multiple factors but also yields comparable accuracy on other compute-bound ML models targeted by most previous methods. Using this performance model and graph-level data and task dependency analysis, we show our system can provide more general model-system co-design than previous methods.
\end{abstract}


\section{Introduction}
\label{sec:intro}
Recommendation models (RMs) have been widely deployed across various industries to improve user experiences and engagements in products and services. Examples include search~\cite{Tennenholtz:2019:RSE}, shopping~\cite{Smith:2017:TDO}, media consumption~\cite{Covington:2016:DNN, Elahi:2020:LRO}, and social networking~\cite{Song:2020:TAN}. Driven by ever-increasing demands, training these models for better prediction rates has become both data- and computationally intensive by involving training data with hundreds of billions of samples, model sizes of up to multiple TBs~\cite{Zhao:2019:ACP}, and multiple (often hundreds of) hosts and devices~\cite{Zhao:2020:DHG} for distributed training. This situation incurs high resource demands for development, debugging, and optimization, which significantly affects the productivity of ML engineers and operation cost of data centers. Therefore, a performance model that accurately predicts an RM’s training performance (e.g., speed, memory usage, etc.) based on its configurations (e.g., batch size, data sharding, number of layers, etc.) is very useful. It removes dependencies on hardware for some tasks and relieves these resource burdens. The flexibility to get performance metrics for varying inputs and configurations helps researchers answer what-if questions, identify bottlenecks, and better meet design constraints. Example questions that performance models can help to answer include but are not limited to: 1) how does changing batch size and/or number of parameters impact performance and memory constraints; 2) how much performance can be gained with new GPUs; 3) can optimizations such as operator (op) fusion improve performance; 4) how to improve embedding table sharding load balance, etc. However, building such a performance model faces three major challenges:

\begin{figure}
  \begin{center}
    \centerline{\includegraphics[width=\columnwidth]{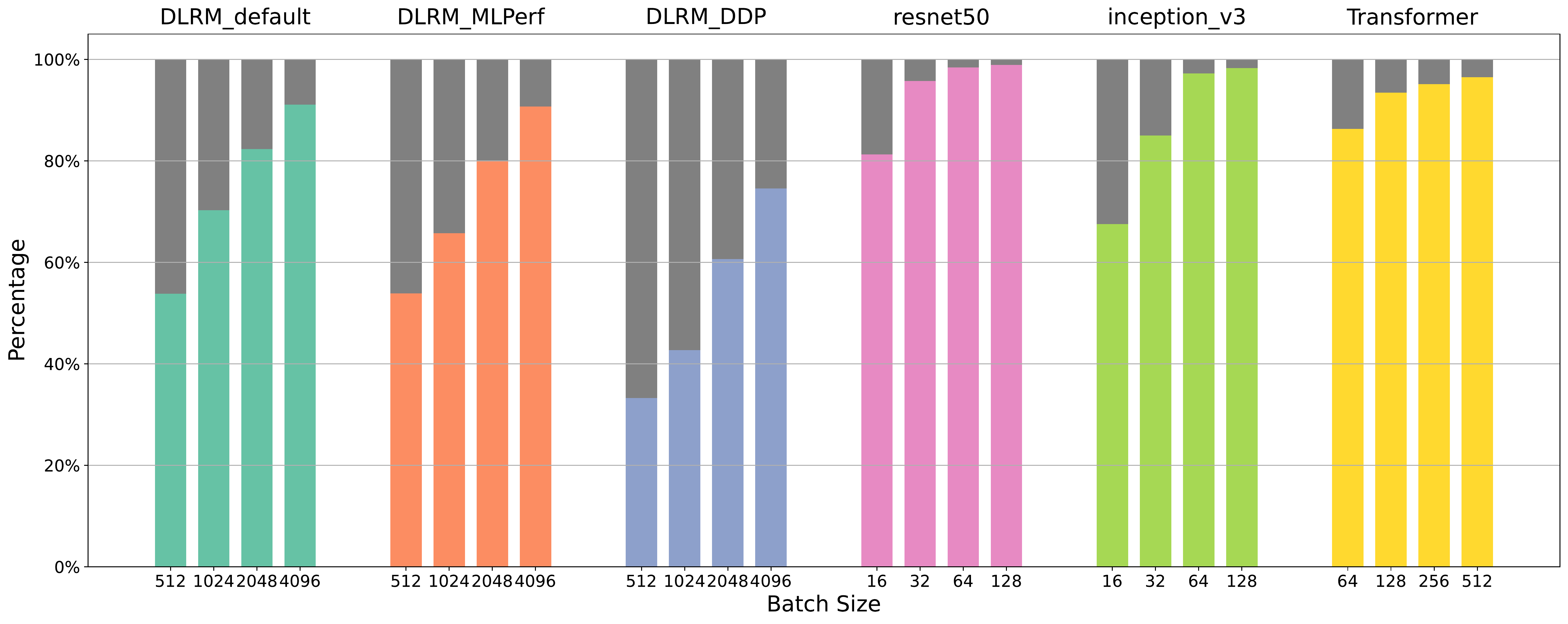}}
    \vskip -0.15in
    \caption{GPU utilization of per-batch training time of six DL models on a NVIDIA Tesla V100 GPU\@. Batch sizes shown here are those commonly used in training. RMs such as \op{DLRM\**} have substantially more device idle time than other models. Whereas other models can be adequately modeled by summing kernel time, modeling RMs is a more complex problem. }
    \label{fig:gpu_util}
    \vskip -0.35in
  \end{center}
\end{figure}

\begin{itemize}
  \item Models with lower GPU utilization are difficult to model. We quantify ``GPU utilization'' as the ratio of \emph{GPU active time} (i.e., when kernels for compute or data transfer are running on the device) over \emph{total training time} per batch.\footnotemark Figure~\ref{fig:gpu_util} shows that the GPU utilizations of some vision (CV) and natural language processing (NLP) models like ResNet~\cite{He:2015:DRL} and Transformer~\cite{Vaswani:2017:AIA} are close to 100\%, whereas that of RMs (with DLRM~\cite{Naumov:2019:DLR} as an example) are much lower. While end-to-end (E2E) runtime of workloads with high GPU utilization can be accurately modeled by simply adding their constituent kernel runtimes, the same method fails for workloads with lower GPU utilization like RMs. \footnotetext{Slightly different from \emph{nvidia-smi}'s definition of GPU utilization (measured over a sample period between 1 and 1/6 second). Notice that ``GPU utilization'' here is a temporal metric and should be distinguished from hardware utilization.}
  \item The combination of GPU asynchronous execution with task/data dependencies makes it difficult to estimate the contribution of each operator's device kernel time and host-side overhead to the per-batch training time on the device.  Previous approaches focused on op-level execution times did not account for these complexities, missing opportunities for a more general and accurate approach.
  \item Finally, an RM comprises a broader range of operators than convolution-dominated CNNs and matrix-multiply-dominated Transformers. While simple models with one kernel performance model may suffice for these simpler cases, RMs require more kernel performance models to characterize their behavior.
\end{itemize}
In summary, previous performance models for DL workloads were not accurate enough to model DLRM and by extension other complex workloads because they did not address low-GPU-utilization, asynchronous, or many-complex-operator workloads.

Our research\footnote{Code is open-sourced at \url{https://github.com/owensgroup/ml_perf_model}.} addresses these complexities by proposing a new performance model for the GPU training of DLRM\@. Once built, this performance model can provide high-confidence metrics to answer questions proposed above and beyond without the need to profile \textit{new} workloads on GPUs.  Here, we focus on a single-GPU configuration in the context of the above challenges, leaving multi-GPU for future work. We begin by analyzing the device execution time of DLRM to identify dominating operators and kernels. Then, using heuristic or ML approaches, we build performance models for these kernels for a wide range of input configurations, and achieve less than 10\% geometric mean average error (GMAE) for each of them in predicting kernel execution time. Beyond accurate kernel models, we also must incorporate host-side overheads into our model. This analysis is a key insight of our work. We categorize host-side overheads into five types and experimentally show that these overheads are consistent across different ops. Using our runtime observer inside PyTorch, we record DLRM's execution graph for its inputs, outputs, and data dependencies. Combining the above components and the ML model execution graph, we construct a critical-path-based E2E performance model for DLRM training on GPUs. This method achieved 4.61\% and 7.96\% geomean errors for per-batch GPU active time and training latency, respectively, compared to actual measured time collected by running the DLRM benchmark. We demonstrate that using shared overheads across workloads only incurs a slight 2.19\% prediction error increase compared to using individual workloads' overheads. This means a user can maintain a shared database for large-scale predictions for numerous workloads. We compared our performance model with several existing performance models on representative CV and NLP models beyond DLRM\@. The results show that our method is general and works well across a variety of workloads on different generations of GPUs.  We also discuss potential use cases of our performance model at the end of the paper, where we demonstrate the model's ability to provide insights into the RM workload characterization and assist practical model-system co-design with the support of the execution graph.  Our contributions in this research include:
\begin{itemize}
  \item For predicting GPU training time of DL models, we show our critical-path-based E2E performance model is a more generalized solution than previous methods that only focus on the device active time, especially those with low GPU utilization such as DLRM\@.
  \item We separately predict kernel time and GPU idle time and show that compared to op-based methods,  this separation facilitates performance modeling by \emph{sharing kernel performance models across ops} that call the same type of kernels and thus reducing the cost of collecting metrics from microbenchmarks. The principles and techniques we used to model kernels can model other kernels that are not included in DLRM as well.
  \item With our specialized model execution graph observer that captures data dependencies among ops, we provide more flexible simulation and performance modeling options that together assist model-system co-design than previous methods do. Without actually running the computation on GPUs,  users can model performance impacts optimization of DL models, such as changing batch size, hardware, operator fusion, reordering, and parallelization, by simply transforming and changing the model execution graph.
\end{itemize}

\section{Related Work}
\subsection{Recommendation Models and DLRM}
RMs have evolved from simple regression-based predictive models~\cite{Walker:1967:EOT}, collaborative filtering~\cite{Herlocker2000}, and neighborhood methods~\cite{Ning:2015:ACS} to deep-learning-based RMs~\cite{Guo:2017:DAF,Cheng:2016:WDL,Lian:2018:XCE,Wang:2017:DCN,Naumov:2019:DLR}. Some deep learning models, such as DIEN~\cite{Zhou:2018:DIN}, also consider sequences of users' actions. The key characteristics that differentiate RMs from CNNs and NLPs are a mixture of sparse and dense computations, large training data volumes, and large, potentially unbounded model sizes.

\begin{figure}[h]
  \begin{center}
    \vskip -0.2in
    \centerline{\includegraphics[scale=0.7]{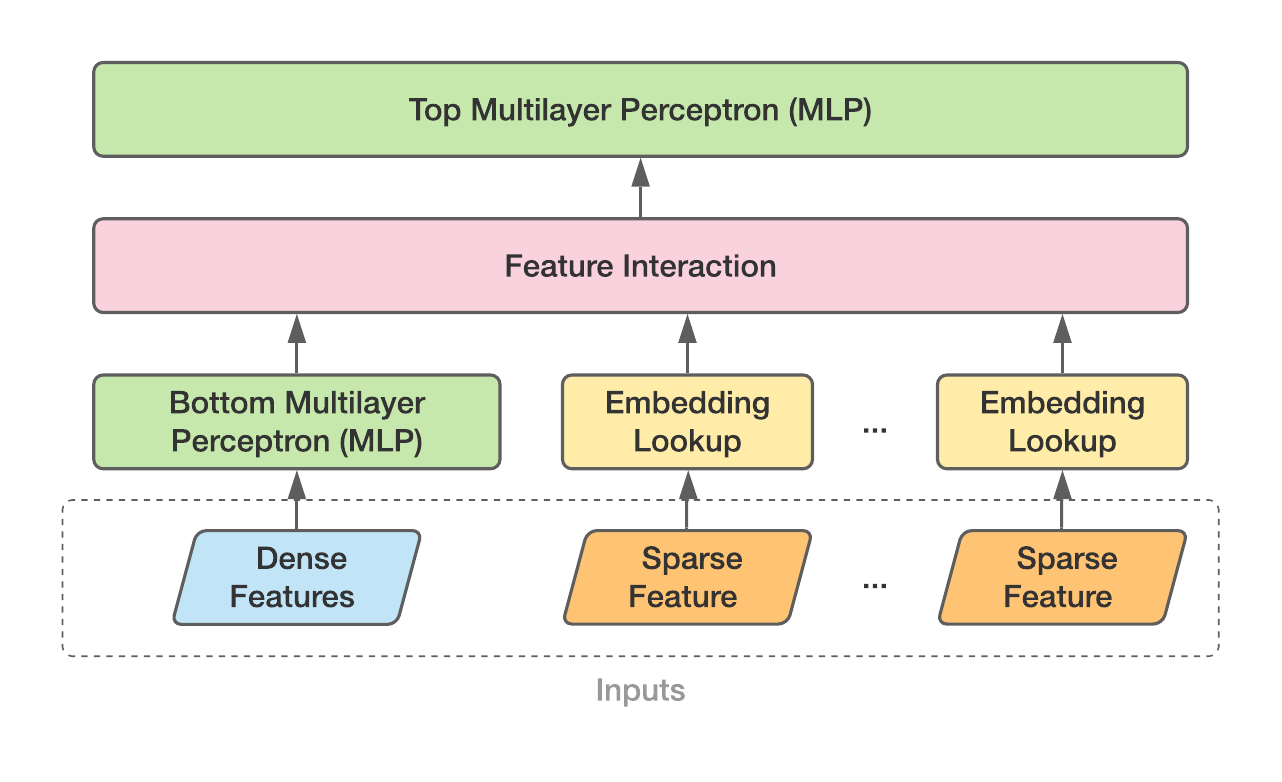}}
    \vskip -0.3in
    \caption{The high-level model architecture of DLRM\@. The inputs (usually user and product data in practice) can be dense and sparse (categorical) features. Each embedding table contains up to millions of embedding vectors and hundreds of values per vector, and because of which they are often sharded across multiple devices in the distributed training.}
    \label{fig:dlrm}
  \end{center}
  \vskip -0.3in
\end{figure}

We choose to use DLRM implemented in PyTorch as a modern representative workload in our analysis. The reasons are: 1) DLRM is a typical example of ML workloads that are highly customizable and at the risk of having low GPU utilization; 2) DLRM forms a common and effective paradigm of using embedding lookup and MLP to process sparse and dense features respectively that generalize to RM design. Figure~\ref{fig:dlrm} depicts DLRM's high-level model architecture. In contrast to embedding table lookups, which are memory-intensive, the multilayer perceptron (MLP) operations are compute-intensive, while any or both of them can dominate the execution time. Besides, the feature interaction is bounded by communication if the model is trained on a multi-GPU platform, and the inputs might be memory-capacity-bound if the training data size is large. Compared to other kinds of models, including CNNs and NLPs, DLRM is potentially bounded by these multiple factors, and as a result building a performance model for it is technically more challenging.

\subsection{GPU operator and kernel performance models}
Op-level and kernel-level performance models usually fall into two categories. \emph{Heuristic models} (e.g., the roofline model~\cite{Williams:2009:RAI}) estimate the kernel execution time by estimating memory traffic, floating point operations, etc. \emph{ML-based models} are trained with benchmark data of kernel execution to predict kernel time for any input size.

\paragraph{Models for GEMM-based kernels} With the current PyTorch release, MLP layers (intrinsically matrix multiplication) rely on cuBLAS and its GEMM-based kernels as the low-level implementation on NVIDIA GPUs. Either using the roofline model or designing a heuristic performance model for these kernels turns out to be infeasible because of not only the lack of source code, but also the special tile quantization and wave quantization effects of cuBLAS~\cite{NVIDIA:2021:CDL}. In existing research (e.g., Lym et al.~\cite{Lym:2019:DGP}) on heuristic performance model design for proprietary libraries like cuDNN, many parameters are still opaque or extremely difficult to measure. Therefore, rather than heuristic ones, an ML-based performance model is more suitable in this case. Previous work~\cite{Liao:2020:PPA,Yu:2021:CPP} shows that either a CNN or MLP model is sufficient to capture the performance features of the GEMM operation. In our work, we use MLP to construct the performance model for cuBLAS kernels called by PyTorch ops like \op{addmm}, \op{bmm},  \op{linear}, etc., which are all GEMM-based.

\begin{table}[t]
  \caption{Comparison of our work with previous ones. E2E prediction of Zhu et al.\ is marked as `Limited' as it only estimates the optimization efficacy on certain kernels instead of making predictions for every single kernel.}
  \label{works_comparison}
  \vskip -0.1in
  \centering
  \setlength{\tabcolsep}{4pt}
    \begin{footnotesize}
      \begin{tabular}{lcccr}
        \toprule
        Work & \begin{tabular}{@{}c@{}}Kernel \\ Pred.\end{tabular} & \begin{tabular}{@{}c@{}}Idle Time \\ Pred. \end{tabular} & \begin{tabular}{@{}c@{}}E2E \\ Pred. \end{tabular} & \begin{tabular}{@{}c@{}}Target Model \\ Types \end{tabular} \\
        \midrule
        Justus et al.~\cite{Justus:2018:PTC} & \cmark & \xmark & \cmark & CNNs \\
        Pei et al.~\cite{Pei:2019:ITP} & \cmark & \xmark & \cmark & CNNs \\
        Liao et al.~\cite{Liao:2020:PPA} & \cmark & \xmark & \cmark & CNNs \\
        Zhu et al.~\cite{Zhu:2020:DAE} & \xmark & \xmark & Limited & Multiple \\
        Yu et al.~\cite{Yu:2021:CPP} & \cmark & \xmark & \cmark & Multiple \\
        Rajagopal et al.~\cite{Rajagopal:2021:PAT} & \xmark & \xmark & \cmark & CNNs \\
        Ours & \cmark & \cmark & \cmark & Multiple+RMs \\
        \bottomrule
      \end{tabular}
    \end{footnotesize}
    \vskip -0.25in
\end{table}

\subsection{Model-level performance modeling}
Previous work~\cite{Justus:2018:PTC, Pei:2019:ITP, Li:2020:CAM, Liao:2020:PPA, Yu:2021:CPP, Rajagopal:2021:PAT} mainly focuses on CNNs and/or NLP models, which are primarily dominated by compute-bound convolution or GEMM ops and have high GPU utilization. In contrast, our work targets a more complex model (DLRM) that can be highly customized with multiple dominating factors, and handles DLRM's substantial device idle time in our E2E training time prediction. \emph{Daydream}~\cite{Zhu:2020:DAE} predicts model runtime after certain optimizations by simulating execution based on the kernel-task dependency graph. This work has a similar approach to ours in addressing the timing of both CPU and GPU threads; however, it lacks the ability to directly predict individual kernel runtime. This limits its capability in predictions for varying input and configuration changes without recollecting performance data using hardware.  Separately, \emph{Habitat}~\cite{Yu:2021:CPP} presented a performance predictor using MLP models trained with kernel metrics. It showed that combining \emph{Habitat} and \emph{Daydream} resulted in a higher average error of 16.1\% than \emph{Daydream} alone. We reduce prediction error compared to this previous work by actually predicting the kernel runtime and overheads based on a finer granularity of instrumentation. In addition, \emph{Daydream}'s kernel dependency graph does not capture data dependencies and thus is limited in discovering and predicting the efficacy of other optimizations such as concurrent kernel execution. In our work, data dependencies are well-captured by the execution graph and therefore we can accurately model a wider variety of optimizations, such as performance-model co-design. Table~\ref{works_comparison} summarizes different features implemented in previous work and ours. To the best of our knowledge, our work is the first that can successfully target the performance modeling complexities characteristic of complex models like DLRM\@.

\begin{figure*}[t]
  \begin{center}
    \vskip -0.1in
    \centerline{\includegraphics[width=0.9\textwidth]{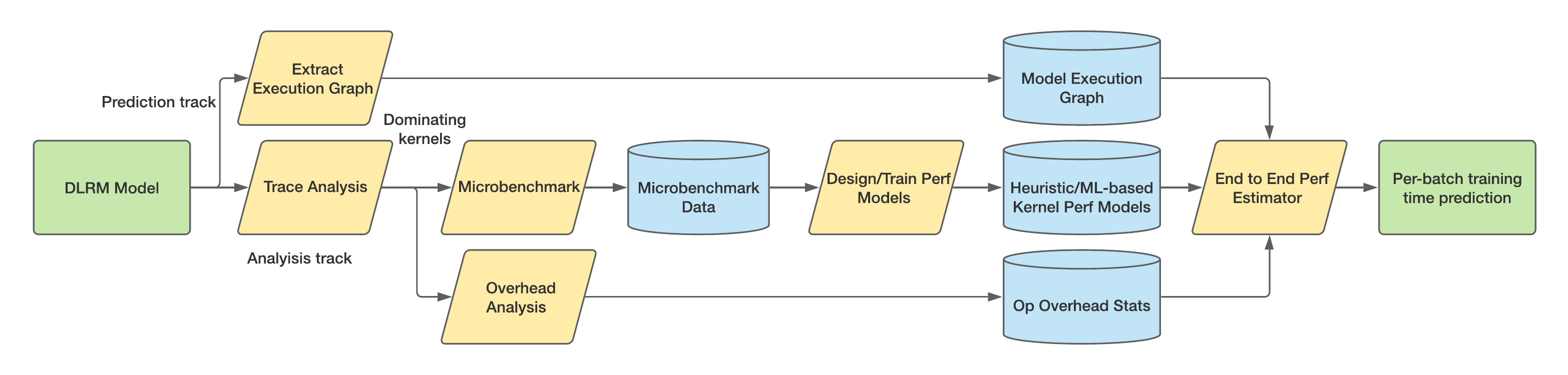}}
    \vskip -0.2in
    \caption{An overview of our prediction pipeline. We begin with DLRM models taken as inputs. These are sent through the \emph{Analysis Track} for trace analysis, microbenchmark data collection, kernel performance model design/training, and op overhead analysis. Armed with these analyses, subsequent DLRM models simply go through the \emph{Prediction Track}, where their execution graphs are extracted and their performance is predicted. This prediction pipeline is designed to be modular so that building blocks of the pipeline marked with blue cylinders can be reused and enriched for modeling tasks for workloads beyond DLRM.}
    \label{fig:system_overview}
  \end{center}
  \vskip -0.3in
\end{figure*}

\section{Methodology}
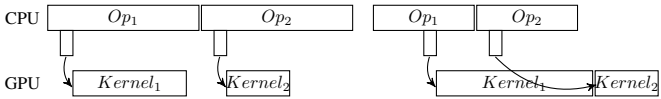
\begin{figure}[h]
  \vskip -0.1in
  \begin{tikzpicture}[scale=0.67, every node/.style={scale=0.67}]
    \node[] at (0,0.25) {CPU};
    \draw[draw=black] (.5,0) rectangle ++(3,0.5) node[pos=.5] {$Op_1$};
    \draw[draw=black] (0.75,-0.5) rectangle ++(0.25,0.5);
    \draw[draw=black] (1,-1.3) rectangle ++(2.25,0.5) node[pos=.5] {$Kernel_1$};
    \node[] at (0,-1.05) {GPU};
    \draw[draw=black] (3.55,0) rectangle ++(3,0.5) node[pos=.5]{$Op_2$};
    \draw[draw=black] (3.8,-0.5) rectangle ++(0.25,0.5);
    \draw[draw=black] (4.05,-1.3) rectangle ++(1.25,0.5) node[pos=.5] {$Kernel_2$};
    \node (A1) at (0.9,-0.4) {};
    \node (B1) at (1.1, -1.25) {};
    \path[->,>=stealth'] (A1) edge[bend right] node [right] {} (B1);
    \node (A2) at (3.95,-0.4) {};
    \node (B2) at (4.15,-1.25) {};
    \path[->,>=stealth'] (A2) edge[bend right] node [right] {} (B2);
  \end{tikzpicture}
  \hfill%
  \begin{tikzpicture}[scale=0.67, every node/.style={scale=0.67}]
    \draw[draw=black] (0,0) rectangle ++(2,0.5) node[pos=.5] {$Op_1$};
    \draw[draw=black] (1,-0.5) rectangle ++(0.25,0.5);
    \draw[draw=black] (1.25,-1.3) rectangle ++(3.1,0.5) node[pos=.5] {$Kernel_1$};
    \node (A1) at (1.15,-0.4) {};
    \node (B1) at (1.35, -1.25) {};
    \path[->,>=stealth'] (A1) edge[bend right] node [right] {} (B1);
    \draw[draw=black] (2.05,0) rectangle ++(2,0.5) node[pos=.5] {$Op_2$};
    \draw[draw=black] (2.3,-0.5) rectangle ++(0.25,0.5);
    \draw[draw=black] (4.4,-1.3) rectangle ++(1.25,0.5) node[pos=.5] {$Kernel_2$};
    \node (A2) at (2.3,-0.4) {};
    \node (B2) at (4.55,-1.05) {};
    \path[->,>=stealth'] (A2) edge[bend right] node [right] {} (B2);
  \end{tikzpicture}
  \caption{Two cases for dependent ops. The small rectangles below the (CPU) ops indicate the launch of their GPU kernels. The left trace is CPU-bound and the right one is GPU-bound. In either case, summing the device active time of the two ops does not properly represent the total execution time, in part because host-side overheads are not considered.}
  \label{fig:synthetic_traces}
  \vskip -0.1in
\end{figure}

Typically, the per-batch training time is estimated by summing the execution time of each \emph{op} in a certain way. Op execution time can be either measured at the host or the device as the sum of kernel execution time.  Since GPU kernels are scheduled asynchronously, it is hard to accurately predict an op's \emph{host time} from the computation it conducts, and thus the op's execution on the device is usually the time to be measured.  For example, CNNs usually resemble the right-hand-side case in Figure~\ref{fig:synthetic_traces}: ops are mostly convolution and GPU compute-bound, and therefore they usually have high GPU utilization. Previous studies that have primarily targeted CNNs can safely make the prediction by summing the individual kernel time and the effects of omitting CPU overheads are minimal. However, this method is sometimes not sufficient to accurately model the E2E execution time, if the model's GPU utilization is low.  As noted in Section~\ref{sec:intro}, DLRM, with its varying sizes and composition of ops, could possibly resemble either the left or right cases in Figure~\ref{fig:synthetic_traces} and have as low as 40\% GPU utilization. This means the per-batch training time prediction error will be 60\% by following the same method, even if the kernel prediction accuracy is 100\%. In practice, execution inefficiencies and inherent model design could both be the cause of low GPU utilization. These complexities necessitate a better methodology of building the performance model for DLRM as well as other models with low GPU utilization.

To address this challenge, we devise a performance modeling pipeline that separates the prediction of device active time and idle time, and integrates both parts with a critical-path-based algorithm that tracks the execution time on both the CPU and GPU\@. Such a separation brings two major advantages in building kernel performance models:
\begin{itemize}
  \item Ops (e.g., \op{addmm}/\op{bmm} vs.\ \op{\{Addmm/Bmm\}Backward}) that have the same type of kernel calls (i.e., cuBLAS GEMM kernels) can \emph{share the same performance model}. This saves us a large amount of time for microbenchmarking and training of ML-based kernel performance models.
  \item Although ML-based performance models can predict kernel time and op overheads as a whole for each op, heuristic models solely based on an op's mathematical expression are not able to address its overheads. Separating them allows us to flexibly choose between these two approaches, while the overheads are handled separately.
\end{itemize}
Figure~\ref{fig:system_overview} depicts an overview of the prediction pipeline. Although we focus on modeling DLRM's performance in this section, it should be noted that this performance model can be handily extended to model ML workloads beyond DLRM by adding new kernel performance models and operator overheads information to the pipeline as assets. Typically, our performance model runs fast and usually finishes a single E2E prediction in a few seconds. The remainder of this section explains how each building block of the pipeline works in detail.

\subsection{Per-batch Training Time Breakdown}
\begin{figure}[b]
  \begin{center}
    \vskip -0.3in
    \centerline{\includegraphics[width=\columnwidth]{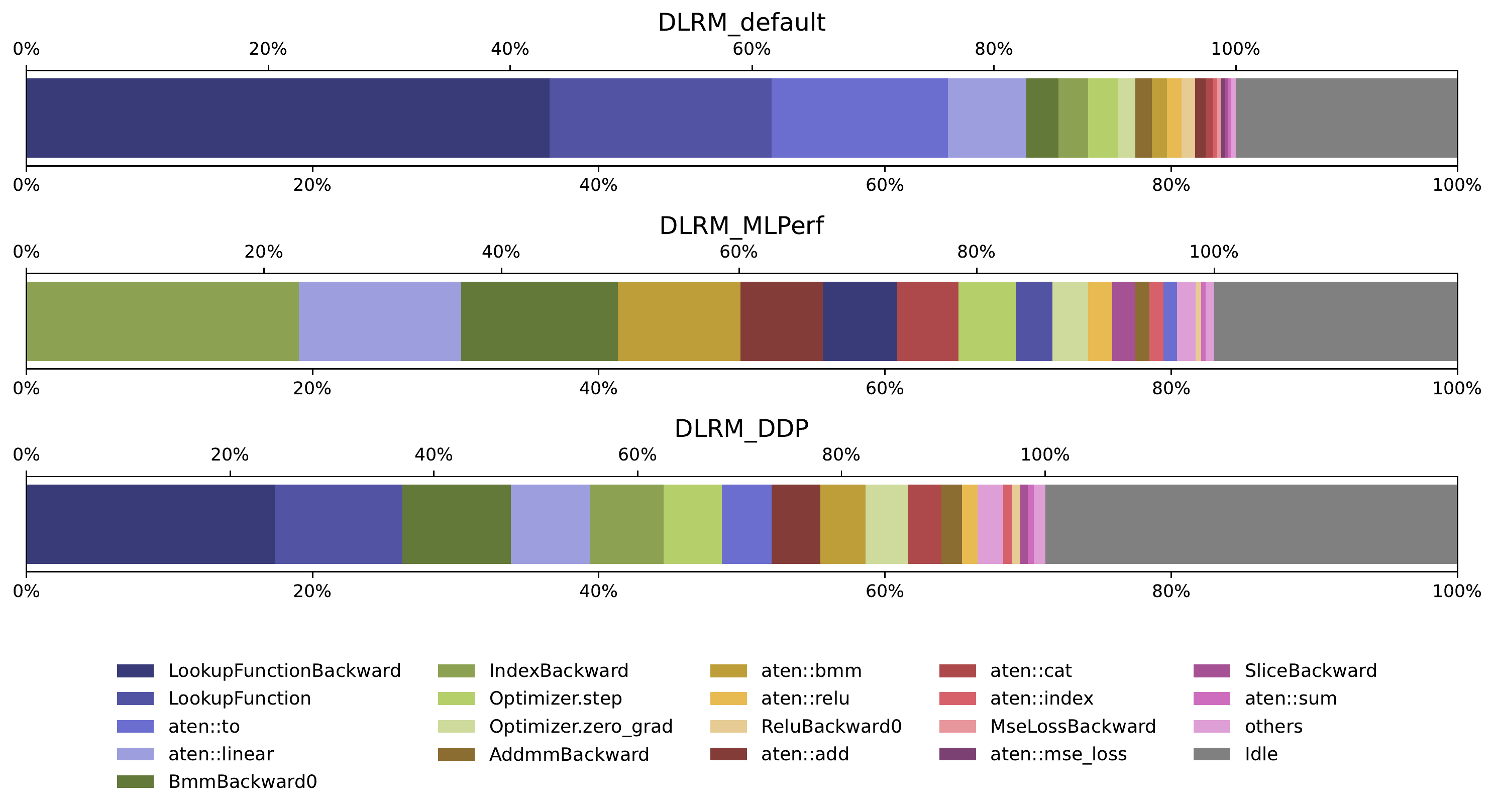}}
    \vskip -0.1in
    \caption{Device time breakdown of three DLRM models with a batch-size of 2048 on a V100 GPU, with profiler overheads \emph{excluded}.  Notice that with different configurations, DLRM is dominated by different kernels, e.g., embedding lookup forward and backward dominates the first and third cases, whereas in the second case it appears to be less important, giving the domination in to \emph{IndexBackward} and FC.}
    \label{fig:active_time_breakdown}
  \end{center}
  \vskip -0.1in
\end{figure}

To understand the device active time and identify dominating ops and kernels, we perform a breakdown of per-batch training time through analyzing PyTorch profiler trace files, in which the metadata of all events, i.e., calls to operators, is flattened. We construct an event tree to represent the calling stack of each op so that the device execution time of each kernel is attributed to the corresponding op, and thus we know the dominating kernels by knowing the dominating ops. The device time breakdown of three DLRM models (configurations shown later) is presented in Figure~\ref{fig:active_time_breakdown}. We observe that:

\begin{itemize}
	\item Just as we noted in Section~\ref{sec:intro},  the device-side idle time forms a non-negligible proportion of the total device time because the host-side op overheads and data dependencies implicitly contribute to it by blocking the scheduling of GPU kernels. This demonstrates the necessity of analyzing kernel execution time and overheads separately.
	\item There is no single op that dominates the device active time of the model. Ops that jointly dominate include compute-bound ops \op{addmm} and \op{bmm}, the memory-bound op \op{embedding lookup}, and communication-bound ops~\op{concat} and \op{to} (memory copy), as well as their counterparts in the backward pass.
	\item Trivial/element-wise ops such as~\op{relu} and~\op{MseLoss} sum to around 5\% of the E2E time. This means they should not be omitted in order to achieve high prediction accuracy.
\end{itemize}

Furthermore, we perform an in-depth analysis on the kernel composition of the dominating ops. The analysis reveals that most of them are composed of or dominated by one single kernel. Exceptions include\op{AddmmBackward} and \op{BmmBackward0} that are dominated by two GEMM kernels, and \op{Optimizer}'s forward and backward ops that are both dominated by a series of element-wise kernels.  Ops in the last category are handled by predicting their sum of kernel time as a whole, possibly ignoring minor kernels that do not appreciably impact the run time. We conclude that there are six major kernels that dominate the per-batch device active time for DLRM training: sparse embedding lookup kernels (both forward/backward) for embedding table lookup, GEMM kernels for bottom and top MLP, and four memory kernels including concatenation, data copy, tensor permutation, and \op{IndexBackward} (low triangular matrix extraction and flatten in feature interaction).

\subsection{Microbenchmark and Performance Models for Dominating Kernels in DLRM}
We create microbenchmarks for seven kernels in total based on the results we get from the breakdown: the six mentioned above plus the trivial \op{IndexForward} that partners with \op{IndexBackward}.  We run the microbenchmarks that sweep through a wide range of (up to 30k) tensor shapes and arguments for each target kernel and take days to run. Specifically, since all GEMM-related ops are dominated by one or two GEMM kernel calls, we skip benchmarking all these ops and share the GEMM kernel benchmark data for their performance modeling. We also discover that the only one type of tensor permutations that occurs in DLRM is the batched matrix transpose, i.e., permutation of the second and third axes of a 3D tensor, and thus it becomes the only type of permutation we benchmark. We first execute the corresponding PyTorch operators on one single GPU for 5 iterations as warm-up, then use NVIDIA's \emph{nvprof} profiler to extract the name of the dominating kernels, and then solely benchmark these kernels for 30 iterations to extract their execution time. Default GPU application clocks are applied, and the CPUs' turbo boost is turned off to guarantee both the accuracy and stability of the benchmark.

With this data, we are able to develop kernel performance models for each of the dominating kernels, as it is impossible to apply one single such model to accurately predict the kernel execution time for all dominating kernels we identify. These performance models are designed in two different ways:
\begin{enumerate}
\item For kernels without \emph{source code access}, such as cuBLAS, PyTorch JIT generated kernels, etc., we predict their execution time with ML-based performance models trained and verified with microbenchmark data.
\item For kernels that are either \emph{accessible or trivial}, i.e., element-wise, we predict their execution time by either using the roofline model or designing heuristic performance models with memory and throughput estimation through code analysis. As such, the microbenchmark data is solely used to verify the prediction accuracy.
\end{enumerate}

The following subsections elaborate how these kernel performance models are developed. Our performance models are highly extensible, as the principles and techniques we introduce (code analysis, ML-based kernel performance model training, etc) also apply to any new ops not covered by this work.

\subsubsection{Heuristic Performance Models}
\paragraph{Characterizing the Embedding Lookup Kernels}
The embedding lookup layers are intrinsically SpMM operations that map categorical features to dense representations. Therefore, the procedure that we describe here for modeling embedding lookup kernels also apply to other kernels of a similar type with irregular memory access patterns and/or is possibly bound by GPU global memory bandwidth. Given a matrix of vector of weights $A \in \mathbb{R}^{m \times t}$ that contains $t$ multi-hot vectors of length $m$ and an embedding table (weight matrix) $W \in \mathbb{R}^{E \times d}$, the embedding lookup operation can be written as $S = A^TW$. Since a real industrial-scale DLRM model usually contains multiple embedding tables, we can simply concatenate these embedding tables, and pack and batch the input indices into new input tensors, such that the embedding lookup operation over multiple embedding tables can be done in one pass. We integrate the implementation of this batched embedding table lookup algorithm (with SGD for the backward case) from Tulloch~\cite{Tulloch:2020:BEL} into DLRM\@. The following analysis is based on the code of this implementation. Important parameters of the implementation include $B$ as the batch size, $E$ as the number of embeddings per table, $T$ as the number of tables, $L$ as the number of lookup operations to produce one dense vector, and $D$ as the embedding vector length. Note that we extend the definition of ``warp'' for simplicity and refer to a group of threads that all have the same $\textit{blockIdx}.x/y/z$ and $\textit{threadIdx}.y/z$ as a \warpsc. In practice, this typically refers to groups of threads of sizes 32, 64, or 128.

We spot that the bounding factor of this op is the memory traffic caused by looking up embedding vectors from the weight tensor. In practice, the value of $E$ can range from a few hundreds to thousands of millions, while $L$ is much smaller, i.e., up to one hundred. We can expect that embedding vectors are more frequently fetched from DRAM than from L2 cache. Therefore, we approximate the execution time of the forward kernel by its DRAM access time, which is given by
\begin{align*}
  \textit{tr\_table\_offsets}_w &= 32~\textrm{bytes}\\
   \textit{tr\_offsets}_w &= 64~\textrm{bytes}\\
  \textit{tr\_indices}_w &= \lceil 4 \times L / 32\rceil \times 32~\textrm{bytes} \\
  \textit{tr\_weights}_w = \textit{tr\_outputs}_w &= \lceil 4 \times D / 32\rceil \times 32~\textrm{bytes} \\
  t &= \frac{\textit{DRAM\_traffic}}{\textit{peak\_DRAM\_BW}} \\
    &= \frac{B \times T \times (\textit{sum of all above})}{\textit{peak\_DRAM\_BW}}.
\end{align*}

The subscript $w$ denotes that these are per-\warpsc\ DRAM traffic; $B \times T$ is the total number of \warpsc{}s. For the backward kernel, we simply replace the per-\warpsc\ weights traffic by \[\textit{tr\_weights}_w = \lceil 2 \times 4 \times L \times D / 32\rceil \times 32~\text{bytes},\] and follow exactly the same other equations.

This method can be further enhanced by estimating the L2 cache hit rate of accessing the embedding lookup table and separating the total memory traffic into DRAM traffic and L2 traffic. As one thread \warpsc\ is responsible for computing one vector in the output tensor, assuming only one CTA resides on each streaming-multiprocessor (SM) on the GPU at a time, the number of embedding lookup tables whose (at least part of) data simultaneously reside in L2 cache is given by \[\textit{num\_tables} = \textit{rows\_per\_block} \times (\# \textit{SM}) / B,\] where $\textit{rows\_per\_block}$ is a kernel argument specifying how many output vectors are computed per CTA\@. With the L2 cache size of the GPU known to us, we can calculate the number of rows per table that resides in the L2 cache as \[\textit{avg\_cached\_rows\_per\_table}=\min\left(\frac{\textit{L2\_cache\_size}}{(\textit{num\_tables}) \times D}, E\right),\] where the second term covers the case when an embedding lookup table with $E$ rows is small enough to reside in the L2 cache. Therefore, the hit rate of the L2 cache, i.e., the probability that the accesses to a total of $L$ embedding lookup table row vectors among all $E$ vectors, can be estimated by \[p = \frac{{\textit{avg\_cached\_rows\_per\_table} \choose L}}{{E \choose L}}.\]

Notice that the $\textit{table\_offsets}$ and $\textit{offsets}$ tensors are relatively very small and frequently accessed, and thus we assume they always stay in L2. Therefore, we construct the enhanced performance model as:
\begin{align*}
  \textit{tr}_\textit{L2} = ~& \textit{tr\_table\_offsets}_w + \textit{tr\_offsets}_w + p \times \textit{tr\_weights}_w \\
  \textit{tr}_\textit{DRAM} =~& \textit{tr\_indices}_w + \textit{tr\_outputs}_w + (1-p) \times \textit{tr\_weights}_w \\
  t =~& \frac{\textit{DRAM\_traffic}}{\textit{peak\_DRAM\_BW}} + \frac{\textit{L2\_traffic}}{\textit{peak\_L2\_BW}} \\
    =~& B \times T \times \left(\frac{\textit{tr}_\textit{DRAM}}{\textit{peak\_DRAM\_BW}} + \frac{\textit{tr}_\textit{L2}}{\textit{peak\_L2\_BW}}\right).
\end{align*}

\paragraph{Characterizing Element-wise Kernels}
For memory kernels of ops including \emph{concat}, \emph{memcpy}, etc.\ that involve intra-GPU or CPU-GPU data transfer, as well as element-wise kernels of ops like \emph{ReLU}, \emph{sigmoid}, etc., it is straightforward to estimate their execution time by applying the roofline model~\cite{Williams:2009:RAI}:
\begin{align*}
t &= \max(\textit{t\_compute}, \textit{t\_memory}) \\
     &= \max\left(\frac{\textit{FLOP}}{\textit{peak\_throughput}}, \frac{\textit{bytes}_\textit{read} + \textit{bytes}_\textit{write}}{\textit{peak\_BW}}\right).
\end{align*}
We use the maximum measured bandwidth of the benchmark as the corrected peak bandwidth in calculation.

\subsubsection{ML-based Performance Models}
Dominating kernels of DLRM that require ML-based performance models include GEMM, transpose, and the forward and backward kernels of tril, for their source code being either non-accessible or too complex to model heuristically. Specifically, we find that it is non-trivial to model the performance of transpose ops like \emph{T} or \emph{permute}, because technically the underlying implementations of tensor transpose might differ significantly~\cite{Gomez:2016:IPM,Vedurada:2018:TAE}, yet these implementations are opaque to users in PyTorch since the kernel is JIT-generated. Therefore, we adopt the ML performance modeling approach for transpose kernels.

\begin{table}[h]
  \vskip -0.15in
  \caption{MLP performance model search space.}
  \label{table:search_space}
  \vskip -0.1in
  	\centering
        \setlength{\tabcolsep}{1pt}
    \begin{footnotesize}
      \begin{tabular}{lc}
        \toprule
        Hyperparameter & Range \\
        \midrule
        num\_layers & [3,4,5,6,7]\\
        num\_neurons\_per\_layer & [128,256,512,1024]\\
        optimizer & [Adam,  SGD]\\
        learning\_rate & [1e-4, 2e-4, 5e-4, 1e-3, 2e-3, 5e-3, 1e-2] \\
        \bottomrule
      \end{tabular}
    \end{footnotesize}
    \vskip -0.1in
\end{table}

For each kernel in this category, we train a MLP model that takes the kernel's input dimensions as the input features and predict the kernel execution time as the output. We conduct a grid search over a universal search space defined in Table~\ref{table:search_space} for the best configuration by training a series of MLP models over the microbenchmark data and keeping the one with the lowest prediction error. The loss function for training is Mean Square Error (MSE)\@. As the input sizes of the benchmark are chosen in an almost exponential scale, e.g., 32, 64, 128, etc., we preprocess the dataset by taking logarithm values of both the sizes and the results. We also scale the learning rate by 10 if $\textit{SGD}$ is chosen as the optimizer. Typically, obtaining such an MLP model for one kernel through grid search takes a few hours of training on one single GPU\@.

\subsection{Device Idle Time Analysis}
\label{sec:device-idle-time-analysis}
Device idle time, as we show in Figure~\ref{fig:active_time_breakdown}, is an important part of the total device execution time. We predict device idle time based on overheads obtained by analyzing the trace files generated by profilers. In a single-GPU context, the main source of device idle time is the \emph{host overheads that are not hidden}. There are two assumptions we make for these overheads:
\begin{itemize}
  \item Model-independence: Same types of overheads of the same op have the same stats on the same machine.
  \item Size-independence: Overheads do not depend on input/output tensor sizes of ops.
\end{itemize}

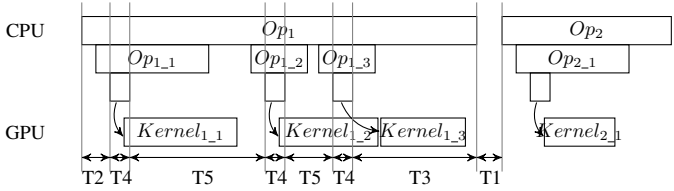
\begin{figure}[t]
  \vskip -0.1in
  \begin{tikzpicture}[scale=0.75, every node/.style={scale=0.75}]
    \node[] at (0,0.25) {CPU};
    \node[] at (0,-1.55) {GPU};

    \draw[draw=black] (1,0) rectangle ++(7,0.5) node[pos=.5] {$Op_1$};
    \draw[draw=black] (8.45,0) rectangle ++(3,0.5) node[pos=.5]{$Op_2$};

    \draw[draw=black] (1.25,-0.5) rectangle ++(2,0.5) node[pos=.5] {$Op_{1\_1}$};
    \draw[draw=black] (1.5,-1) rectangle ++(0.35,0.5);
    \draw[draw=black] (1.75,-1.8) rectangle ++(2,0.5) node[pos=.5] {$Kernel_{1\_1}$};
    \node (A1) at (1.65,-0.9) {};
    \node (B1) at (1.85, -1.75) {};
    \path[->,>=stealth'] (A1) edge[bend right] node [right] {} (B1);

    \draw[draw=black] (4,-0.5) rectangle ++(1,0.5) node[pos=.5] {$Op_{1\_2}$};
    \draw[draw=black] (4.25,-1) rectangle ++(0.35,0.5);
    \draw[draw=black] (4.5,-1.8) rectangle ++(1.75,0.5) node[pos=.5] {$Kernel_{1\_2}$};
    \node (A2) at (4.4,-0.9) {};
    \node (B2) at (4.6, -1.75) {};
    \path[->,>=stealth'] (A2) edge[bend right] node [right] {} (B2);

    \draw[draw=black] (5.2,-0.5) rectangle ++(1,0.5) node[pos=.5] {$Op_{1\_3}$};
    \draw[draw=black] (5.45,-1) rectangle ++(0.35,0.5);
    \draw[draw=black] (6.3,-1.8) rectangle ++(1.5,0.5) node[pos=.5] {$Kernel_{1\_3}$};
    \node (A3) at (5.55,-0.9) {};
    \node (B3) at (6.4,-1.55) {};
    \path[->,>=stealth'] (A3) edge[bend right] node [right] {} (B3);

    \draw[draw=black] (8.7,-0.5) rectangle ++(2,0.5) node[pos=.5] {$Op_{2\_1}$};
    \draw[draw=black] (8.95,-1) rectangle ++(0.35,0.5);
    \draw[draw=black] (9.2,-1.8) rectangle ++(1.25,0.5) node[pos=.5] {$Kernel_{2\_1}$};
    \node (A4) at (9.1,-0.9) {};
    \node (B4) at (9.3,-1.75) {};
    \path[->,>=stealth'] (A4) edge[bend right] node [right] {} (B4);

    \draw[draw=gray] (1,0.75) -- (1,-2.25);
    \draw[draw=gray] (1.5,0.75) -- (1.5,-2.25);
    \draw[draw=gray] (1.85,0.75) -- (1.85,-2.25);
    \draw[draw=gray] (4.25,0.75) -- (4.25,-2.25);
    \draw[draw=gray] (4.6,0.75) -- (4.6,-2.25);
    \draw[draw=gray] (5.45,0.75) -- (5.45,-2.25);
    \draw[draw=gray] (5.8,0.75) -- (5.8,-2.25);
    \draw[draw=gray] (8,0.75) -- (8,-2.25);
    \draw[draw=gray] (8.45,0.75) -- (8.45,-2.25);

    \draw[>=stealth', <->] (8,-2) -- (8.45,-2);
    \node[] at (8.25,-2.35) {T1};

    \draw[>=stealth', <->] (1,-2) -- (1.5,-2);
    \node[] at (1.2,-2.35) {T2};

    \draw[>=stealth', <->] (5.8,-2) -- (8,-2);
    \node[] at (7,-2.35) {T3};

    \draw[>=stealth', <->] (1.5,-2) -- (1.85,-2);
    \draw[>=stealth', <->] (4.25,-2) -- (4.6,-2);
    \draw[>=stealth', <->] (5.45,-2) -- (5.8,-2);
    \node[] at (1.7,-2.35) {T4};
    \node[] at (4.45,-2.35) {T4};
    \node[] at (5.65,-2.35) {T4};

    \draw[>=stealth', <->] (1.85,-2) -- (4.25,-2);
    \draw[>=stealth', <->] (4.6,-2) -- (5.45,-2);
    \node[] at (3.15,-2.35) {T5};
    \node[] at (5.05,-2.35) {T5};
  \end{tikzpicture}
  \vskip -0.15in
  \caption{Host-side overhead types. The labels T1--T5 indicate the five overhead types introduced in Section~\ref{sec:device-idle-time-analysis}. Each op has one T2 and one T3 overhead, and at least one T4 overhead if it has device kernel calls. }
  \label{fig:overheads}
\end{figure}

That means overheads are supposed to be only dependent to the training platform (i.e., CPUs) configurations. Based on these two assumptions, we analyze the host-side overheads and categorize them into five types as shown in Figure~\ref{fig:overheads}, including:
\begin{itemize}
  \item Type 1: Overhead between two top-level PyTorch op calls.
  \item Type 2: Overhead before an op's first kernel launch begins.
  \item Type 3: Overhead after an op's last kernel launch ends.
  \item Type 4: Execution time of CUDA runtime functions, e.g., \emph{cudaLaunchKernel}, \emph{cudaMemcpyAsync}, etc.
  \item Type 5: Overheads between two kernel launches.
\end{itemize}

Some of the overheads, namely T2, T3, and T5, should be independent of the input parameters of the op, as we assume all the parameter-defined operations, mainly the computation and data movements, are offloaded to the device. By analyzing 100-iteration trace files of the models we choose, we characterize each type of overheads and store their mean values in a JSON file to be used in the E2E performance model. To guarantee the accuracy, profiler overheads of CPU and GPU events are excluded by subtracting them from the execution time of each event. In practice we use 4~$\mu s$ as indicated in the PyTorch source code to model the profiler overheads of GPU events, while that of CPU events vary from platform to platform, and we find that an empirical value of 2~$\mu s$ is a good choice.

\subsection{E2E GPU Training Performance Model}
One challenge of building an end to end performance model of an ML training workload is to have sufficient information about its run-time execution. Early implementations of ML frameworks such as Caffe~\cite{Jia:2014:CCA} define an ML model as a static graph in the protobuf format. In recent years, ML frameworks such as TensorFlow~\cite{Google:2016:TAS} and PyTorch~\cite{Facebook:2019:PAS} have closely integrated programming language bindings to support dynamic ML model graphs characterized by conditionals and loops. Furthermore, they support eager mode execution. With the flexibility of these frameworks, the ML model definition is essentially a program and requires execution to fully capture the run-time characteristics. We implemented an execution graph observer inside PyTorch that allows us to extract both the operators executed and their inputs and outputs data dependencies during the model training process. Once the ML model's run-time execution is captured, the execution graph can be reconfigured to use different data inputs or hardware devices. For example, we may collect the execution graph while running on CPU and apply our performance models to the execution graph to predict the workload's performance on the GPU or other types of hardware.

\setlength{\textfloatsep}{0pt}
\begin{algorithm}[t]
  \caption{E2E GPU Training Performance Model.}
  \label{alg:perf_model}
  \begin{algorithmic}[1]
    \STATE {\bfseries Input:} Execution graph $G$ of a DLRM model; Kernel performance models $\{M\}$; Overheads $Ov$.
    \STATE {\bfseries Output:} Predicted per-batch training time $T$.

    \STATE Initialize $cpu\_time = 0, gpu\_time = 0$
        \FOR{each $op$ in $G$}
                \STATE Look up $T1, T2, T3, T4, T5$ from $Ov$ for $op$
                \STATE $cpu\_time \pluseq T1$
                \IF{op has kernel calls}
                        \STATE $cpu\_time \pluseq T2$
                        \FOR{each kernel $k$ $op$ calls}
                                \STATE Predict kernel time $T_k$ with the corresponding performance model picked from $\{M\}$
                                \STATE $gpu\_time = \max(gpu\_time + 1, cpu\_time + T4 / 2) + T_k$
                                \STATE $cpu\_time \pluseq T4$
                                \IF{$k$ is not the last kernel}
                                        \STATE $cpu\_time \pluseq T5$
                                \ENDIF
                        \ENDFOR
                        \STATE $cpu\_time \pluseq T3$
                \ELSE
                        \STATE $cpu\_time \pluseq T5$
        \ENDIF
    \ENDFOR
    \STATE $T = \max(gpu\_time, cpu\_time)$
  \end{algorithmic}
\end{algorithm}

We devise a critical-path-based Algorithm~\ref{alg:perf_model} that integrates the predicted kernel time and overheads to predict the E2E training time of DLRM\@. We identify the critical path of execution by keeping track of both the execution time on CPU and GPU\@. For each operator, we first add T1 and T2 to the CPU time as a prerequisite. If the op has kernel calls, we set the start time of each kernel based on whether the CPU or GPU time is the critical path (line 11), so that the device idle time caused by the host overheads is counted. Each kernel time is then added to the GPU time, while T4 and T5 are added to the CPU time. T3 is added after all kernels are processed. Eventually, we take the maximum of CPU and GPU time as the critical-path and thus the final E2E predicted time.

\section{Results and Analysis}
We evaluate our benchmark and performance models on three different NVIDIA GPUs---Tesla V100, Tesla P100, and GeForce GTX TITAN Xp---with CUDA 11.3 and Python 3.9. We conduct the E2E tests on three open-sourced DLRM models that can be accessed in Meta's DLRM repo on Github~\cite{Facebook:2019:DGR}. We name them DLRM\_default, DLRM\_MLPerf, and DLRM\_DDP, and show their configurations in Table~\ref{DLRMs}. To launch the training of the DLRM\_MLPerf model, we use the Kaggle Criteo dataset as the training dataset, and change the embedding table sparse feature size of \op{DLRM\_MLPerf} from 128 to 32 to allow it to fit into the memory of our TITAN Xp and P100\@. We also use the code repository of Konstantinidis et al.~\cite{Konstantinidis:2017:AQR} to benchmark the GPU hardware parameters, e.g., FLOPS, DRAM bandwidth, etc., that are needed by the heuristic performance models.

\begin{table}[h]
  \vskip -0.1in
  \caption{DLRM model configurations}
  \label{DLRMs}
  \vskip -0.1in
  \setlength{\tabcolsep}{1pt}
  \centering
    \begin{footnotesize}
      \begin{tabular}{@{}lccc@{}}
        \toprule
         & DLRM\_default & DLRM\_MLPerf & DLRM\_DDP \\
        \midrule
        Bot MLP & 512-512-64 & 13-512-256-128 & 128-128-128-128\\
	    EL Tables & 8 & 26 & 8\\
	    Rows & 1000000 & Up to 14M & 80000\\
	    EL Dim & 64 & 128 & 128\\
        Top MLP & 1024-1024-1024-1 & 1024-1024-512-256-1 & 512-512-512-256-1\\
        \bottomrule
      \end{tabular}
    \end{footnotesize}
    \vskip -0.15in
\end{table}

\begin{table*}[t]
  \caption{Execution time prediction error for each of the dominating kernels. Abbreviation examples: \textbf{EL} (embedding lookup), \textbf{GEMM} (fully connected and interaction layers),  \textbf{memcpy} (memory copy from host to device),  \textbf{concat} (concatenation),  \textbf{tril} (lower triangular extraction and flatten),  \textbf{F} (forward), \textbf{B} (backward), \textbf{H} (with hit rate estimation for EL),  \textbf{L} (large size,  average embedding table size greater than 100000).}
  \label{kernel_prediction_error}
  \vskip -0.1in
  \centering
    \begin{small}
      \begin{tabular*}{0.88\textwidth}{lc|ccc|ccc|cccr}
        \toprule
        \multirow{2}{*}{Approach} & GPU & \multicolumn{3}{c}{V100} & \multicolumn{3}{c}{TITAN Xp} & \multicolumn{3}{c}{P100} \\
                                                    & Kernel & GMAE & mean & std & GMAE & mean & std & GMAE & mean & std\\
        \midrule
        		  \multirow{10}{*}{Heuristic} &
                EL-F & 11.46\% & 35.92\% & 56.81\%    & 12.81\% & 34.05\% & 38.92\%       & 8.63\% & 33.19\% & 54.72\% \\
                &EL-FL  & 6.93\% & 11.22\% & 8.96\%    & 7.54\% & 16.76\% & 16.01\%       & 2.89\% & 5.52\% & 6.26\% \\
                &EL-FH  & 9.27\% & 16.73\% & 16.39\%    & 11.88\% & 25.44\% & 26.04\%       & 6.42\% & 13.06\% & 14.81\% \\
                &EL-FHL & 7.85\% & 12.68\% & 10.02\%    & 8.84\% & 18.20\% & 16.68\%       & 3.84\% & 7.02\% & 7.08\% \\
                &EL-B   & 9.53\% & 34.39\% & 60.91\%    & 8.31\% & 38.62\% & 65.77\%       & 12.49\% & 35.26\% & 62.70\% \\
                &EL-BL  & 5.27\% & 5.94\% & 2.29\%    & 2.38\% & 2.95\% & 1.61\%       & 9.88\% & 10.13\% & 2.37\% \\
                &EL-BH  & 7.39\% & 13.37\% & 15.01\%    & 5.57\% & 15.16\% & 23.99\%       & 8.42\% & 12.59\% & 12.12\% \\
                &EL-BHL & 5.69\% & 6.24\% & 2.28\%    & 2.55\% & 3.21\% & 1.68\%       & 10.19\% & 10.42\% & 2.33\% \\
                \cmidrule{2-11}
                &concat & 5.34\% & 11.45\% & 14.76\%    & 8.17\% & 11.48\% & 9.08\%       & 3.30\% & 6.54\% & 12.63\% \\
                \cmidrule{2-11}
                &memcpy & 0.57\% & 0.96\% & 2.46\%    & 7.05\% & 13.87\% & 17.45\%       & 5.10\% & 7.95\% & 8.28\% \\
                \midrule
                \multirow{4}{*}{ML-based} &
                GEMM   & 5.80\% & 10.00\% & 10.33\%    & 8.92\% & 14.24\% & 11.83\%       & 7.59\% & 12.30\% & 10.39\% \\
                \cmidrule{2-11}
                &transpose & 2.95\% & 5.47\% & 6.71\%    & 5.75\% & 10.13\% & 9.67\%       & 3.35\% & 5.92\% & 6.84\% \\
                \cmidrule{2-11}
                &tril-F & 2.13\% & 3.67\% & 3.81\%    & 3.23\% & 6.54\% & 8.17\%       & 3.71\% & 6.74\% & 8.31\% \\
                &tril-B & 3.67\% & 7.35\% & 9.40\%    & 3.08\% & 6.69\% & 9.30\%       & 2.71\% & 4.76\% & 4.51\% \\
        \bottomrule
      \end{tabular*}
    \end{small}
    \vskip -0.25in
\end{table*}

\subsection{Performance Models for Dominating Kernels in DLRM}
In Table~\ref{kernel_prediction_error} we can see that on all types of GPU, our plain performance model for batched embedding table lookup achieves a varying yet low error rate for all table sizes and a stable and lower error rate for big table sizes ($E > \text{100k}$). This is because when the lookup tables are small, the L2 cache can capture substantial locality, and thus our assumption that lookup traffic comes from DRAM is no longer valid. However, with our enhanced performance model, we successfully reduce and stabilize the error rate for all table sizes while still maintaining a lower error rate for big table sizes. Thus we adopt the enhanced model in our E2E analysis. Except for embedding lookup, we also achieve decent (i.e., less than 10\%) GMAE errors on both ML-based models and other heuristic models for all other kernels. The errors of our kernel performance models correlate across all three different GPU devices.

\begin{figure}[h]
  \begin{center}
  	\vskip -0.15in
    \centerline{\includegraphics[width=\columnwidth]{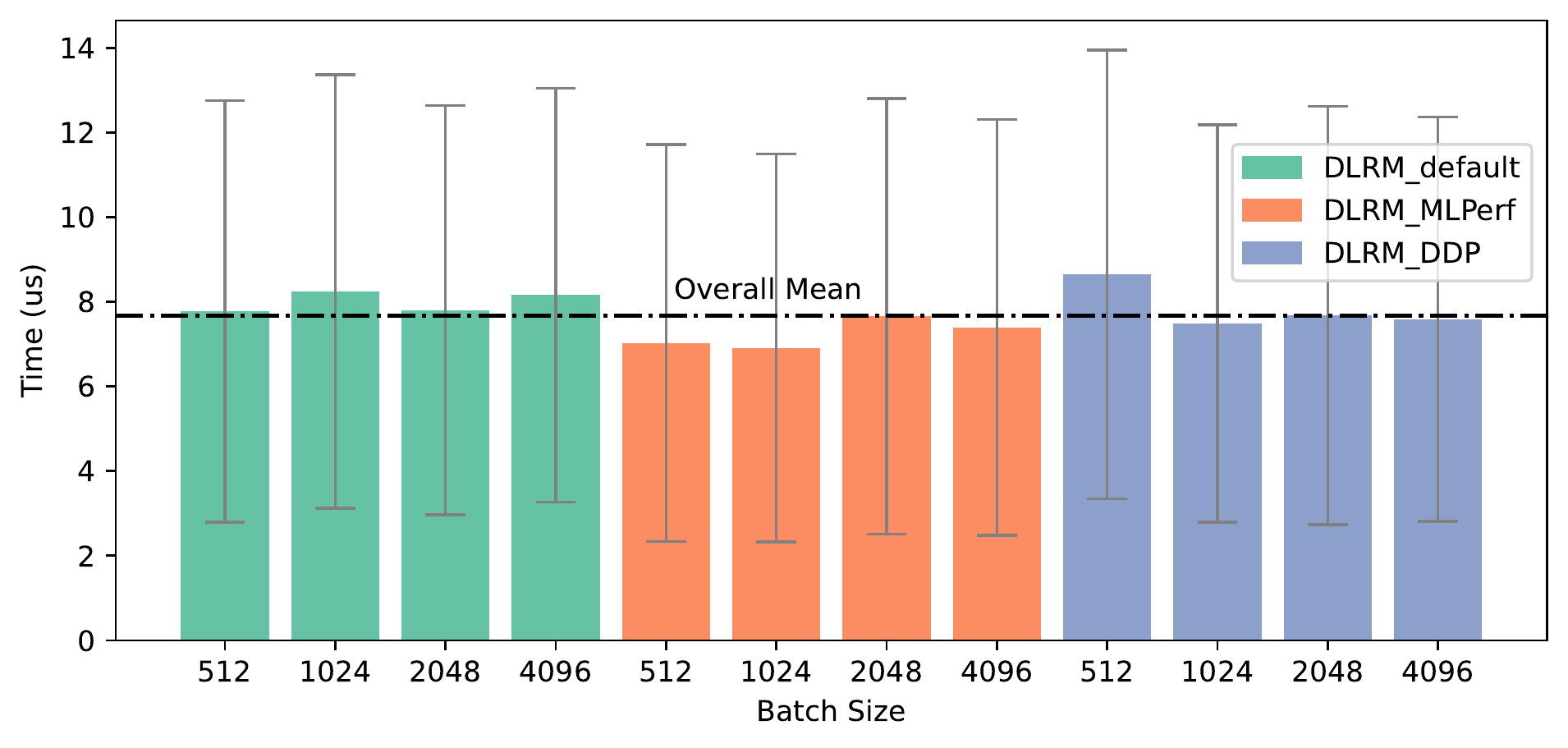}}
    \vskip -0.2in
    \caption{T1 overhead mean and std of all models and batch sizes on V100.}
    \label{fig:t1}
    \vskip -0.4in
  \end{center}
\end{figure}

\subsection{Overheads Analysis}
\begin{figure*}[t]
  \begin{center}
    \vskip -0.15in
    \centerline{\includegraphics[width=0.9\textwidth]{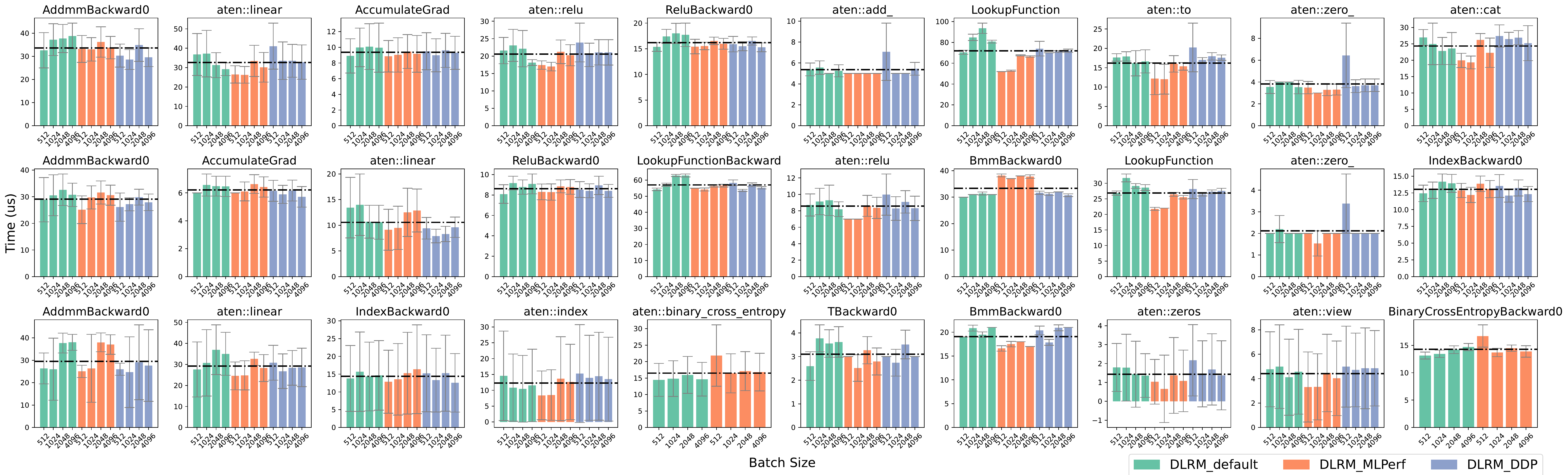}}
    \vskip -0.15in
    \caption{Overhead mean and std of 10 most dominating ops per overhead type for each models and batch size on V100. Each row represents T2, T3, T5, respectively in top-down order. Overall means of each overhead type per op are plotted in dash lines in each subplot.}
    \label{fig:t235}
  \end{center}
  \vskip -0.2in
\end{figure*}

We perform analysis on overheads extracted from collected traces of models' E2E execution. We remove per-type outliers outside whiskers ($Q1-1.5\textit{IQR}, Q3+1.5\textit{IQR}$) for each individual workloads. The reason we do not conduct an op-level microbenchmark for overheads is because it hardly simulates the overhead behaviors in actual E2E execution. Fig.~\ref{fig:t1} and~\ref{fig:t235} show the statistics of T1 and T2/3/5 overheads respectively. We omit T4 here as we use a value of 10~$\mu s$ to approximate all the CUDA runtime functions. We see that the means of T1 of different models and batch sizes are close to each other around 8~$\mu s$. With different overall mean values, the same conclusion holds for all comparison shown in Fig.~\ref{fig:t235}. From these two figures, no trends of model types or tensor sizes (represented by the batch size while treating all ops per E2E run as an ensemble) being able to affect the overhead statistics are observed. Although this is not a strict mathematical proof the model/size-independence, as we only need a simple estimation for the overheads to fill the gap between device active time and per-iteration time, we argue that it is safe to use the mean values of overhead per type per workload in E2E prediction.

\subsection{E2E GPU Training Performance Model for DLRM and More DL Models}
\begin{figure}[h]
  \begin{center}
  	\vskip -0.2in
    \centerline{\includegraphics[width=\columnwidth]{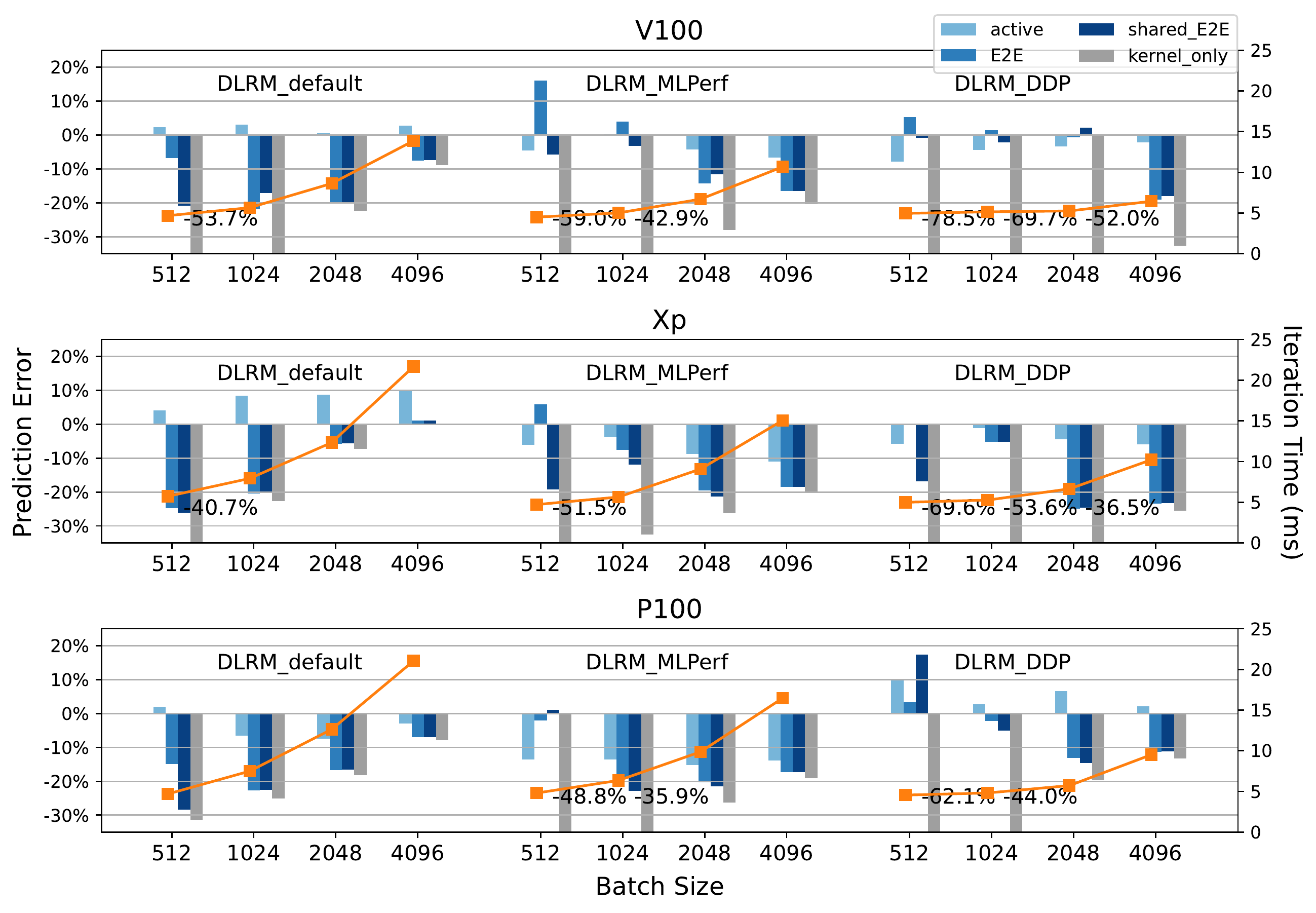}}
    \vskip -0.15in
    \caption{E2E per-batch training time prediction of three DLRM models on three GPUs. \emph{active}, \emph{total}, \emph{kernel\_only} are respectively the prediction errors of GPU active time, E2E per-batch time, and solely using GPU active time without modeled idle time as the E2E time. Measured iteration time is plotted in orange color for reference.}
    \label{fig:e2e}
    \vskip -0.1in
  \end{center}
\end{figure}

We evaluate our E2E prediction on the three DLRM models on three GPUs and show the results in Table~\ref{error_stats} and Figure~\ref{fig:e2e}. The baseline we use ( ``kernel\_only'' in Figure~\ref{fig:e2e}) is the E2E training time prediction error by summing up solely the predicted kernel execution time \emph{without} the modeled overheads i.e.  GPU active time. We predict the E2E training time with our proposed algorithm. Specifically, ``E2E'' means to predict E2E time with overheads from individual workloads, while ``shared\_E2E'' means to predict with \emph{shared overheads aggregated across the workloads}., i.e., averaging the samples across the workloads collected in overhead analysis. We see that the geomean values of active and E2E time prediction error are 4.61\% and 7.96\% respectively.  The E2E prediction error with shared overheads is 10.15\%, only 2.19\% higher than that with individual overheads; this indicates the feasibility of maintaining an overhead database for large-scale ML workload predictions in an industrial environment.  We notice a trend of gaps between \emph{E2E} and \emph{kernel\_only} shrinking as batch size increases. This is because GPU utilization increases with batch size and therefore our performance model degenerates towards ``kernel\_only''. The fact that \emph{kernel\_only} prediction errors are much worse than \emph{E2E} when GPU utilization is low justifies the necessity and success of including the modeling of device idle time in our prediction algorithm.  Device-wise, the GPU active time error on V100 is the lowest among the three, while the E2E error is the lowest on the platform with TITAN Xp. The prediction error of the device active time comes from the kernel execution time prediction error. For example, the MLPerf model has non-constant table sizes and thus we have to use the average table size in the performance model, which affects its accuracy. Overall, the device active time error rate lies within the range of our expectation, proving the success of the kernel performance model. The E2E time predictions have a clear trend of underestimation, which can be explained by the underestimation of device idle time. We suspect that it is because some of the overheads, e.g., T1, or T4 of \op{cudaMemcpyAsync}, etc., have long-tail distributions with high variation, while we remove many upper outliers and use their mean values in the predictions. Since these are usually common overheads (T1 is the most common as it occurs for every single op), the error might accumulate quickly and thus result in underestimation of device idle time and E2E time. In addition, we observe no systematic or correlated errors in either active or idle time and are confident that the E2E device active time and total time are appropriately predicted.

\begin{figure}[h]
  \begin{center}
    \vskip -0.2in
    \centerline{\includegraphics[width=\columnwidth]{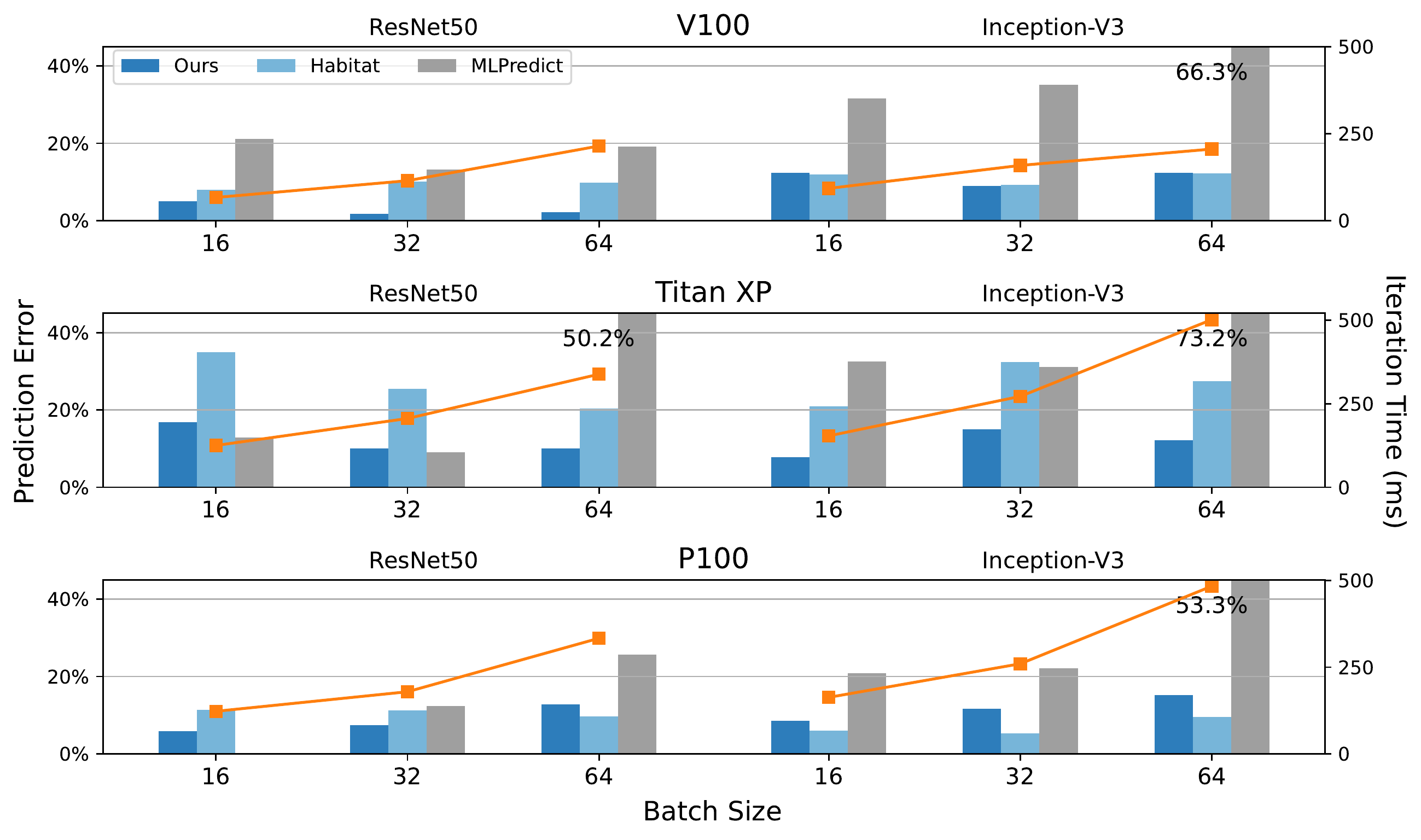}}
    \vskip -0.15in
    \caption{E2E per-batch training time prediction of ResNet50 and Inception-V3 as representatives of non-DLRM DL models on three different GPUs. We used Habitat open source project to collect the prediction result on TITAN Xp since it was not reported in the paper. Actual iteration time is also plotted in orange color for reference.}
    \label{fig:e2e_others}
  \end{center}
  \vskip -0.1in
\end{figure}

As Figure~\ref{fig:e2e_others} shows, we also compare our performance model on two CV models (ResNet50 and Inception-V3) with two previous works, \emph{Habitat}~\cite{Yu:2021:CPP} and \emph{MLPredict}~\cite{Justus:2018:PTC}, neither of which supports DLRM mainly because of their limited coverage of ops. We do not compare with \emph{Daydream} as it is not open-source and does not make E2E predictions. To enable the prediction of these two models, we extend our microbenchmark to cover the convolution and batch-normalization ops. We can see that our work achieves comparable or better prediction errors against the two previous works. The reason \emph{MLPredict} fails to produce accurate results on some tests might be that the pretrained predictor does not cover certain batch sizes (possibly due to GPU memory limits) and/or convolution input sizes (such as Inception-V3's $1\times 7$ and $7\times 1$ convolution filters).

\section{Discussions}
The advantages of our performance model against previous works include: (1) accurate prediction of individual kernel performance and op overheads and (2) op data dependencies capturing with our execution graph. Therefore, we are able answer the questions we ask in Section~\ref{sec:intro} with more comprehensive and flexible performance modeling and simulation options than both previous works and trace file inspection. Typical use cases of our work include iterative model tuning and op optimization such as fusion. Beyond the models and devices used in this paper, our system as shown in Figure~\ref{fig:system_overview} is highly extendible for performance modeling of other types of ML workloads on heterogeneous platforms with types of devices from other vendors such as Intel and AMD\@.

\begin{table*}[t]
  \caption{Statistics of active (kernel) time and E2E time prediction errors across three platforms.}
  \label{error_stats}
  \vskip -0.1in
  \centering
  \setlength{\tabcolsep}{5.5pt}
    \begin{small}
      \begin{tabular*}{\textwidth}{lccc|ccc|ccc|cccr}
        \toprule
        & \multicolumn{3}{c}{\bf{Overall}} & \multicolumn{3}{c}{V100} & \multicolumn{3}{c}{TITAN Xp} & \multicolumn{3}{c}{P100} \\
        & geomean & min & max & geomean & min & max & geomean & min & max& geomean & min & max\\
        \midrule
        Active & \bf4.61\% & 0.41\% & 15.25\% & 2.69\% & 0.41\% & 7.82\% & 5.73\% & 1.18\% & 11.04\% & 6.37\% & 1.99\% & 15.25\%\\
        E2E & \bf7.96\% & 0.09\% & 24.92\% & 7.56\% & 0.73\% & 21.96\% & 6.97\% & 0.09\% & 24.92\% & 9.59\% & 2.04\% & 22.76\%\\
        Shared E2E & \bf10.15\% & 0.75\% & 28.38\% & 6.92\% & 0.75\% & 20.79\% & 12.52\% & 1.13\% & 26.17\% & 12.09\% & 1.06\% & 28.38\%\\
       \bottomrule
      \end{tabular*}
    \end{small}
    \vskip -0.2in
\end{table*}

\subsection{Performance Modeling for Model-System Co-design}
\paragraph{Iterative Model Tuning} To ensure both high precision/recall and fast training speed, the iterative tuning of configurations of ML models (e.g.,  number and size of layers) is necessary yet difficult, especially when frequent training job launches in an industrial environment is costly and not always practical. With our performance model, users can handily make transformations like \emph{insert}, \emph{remove}, \emph{replace}, \emph{resize}, and \emph{parallelize} on our easily mutable execution graph and predict the outcome of their optimization without actually running the code. Specifically, it is straightforward to change metadata of tensor shapes of selected ops and their parent and child nodes in the graph for \emph{resize}, and to assign ops in parallel branches with no data dependency to different GPU streams for \emph{parallel}. This can only be performed with our support of data dependencies between ops and individual kernel runtime prediction. In fact, our performance model could be integrated as a module into network architecture search (NAS) and significantly improve automatic search for the best ML model configuration. We see this as exciting future work.

\begin{figure}
  \vskip -0.1in
  \centering
  \subfloat{\includegraphics[width=.51\columnwidth]{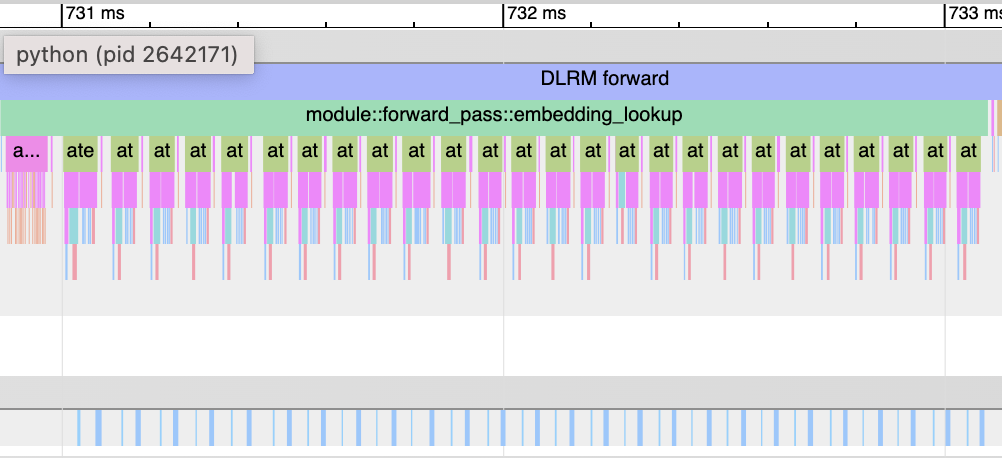}}\hfill
  \subfloat{\includegraphics[width=.43\columnwidth]{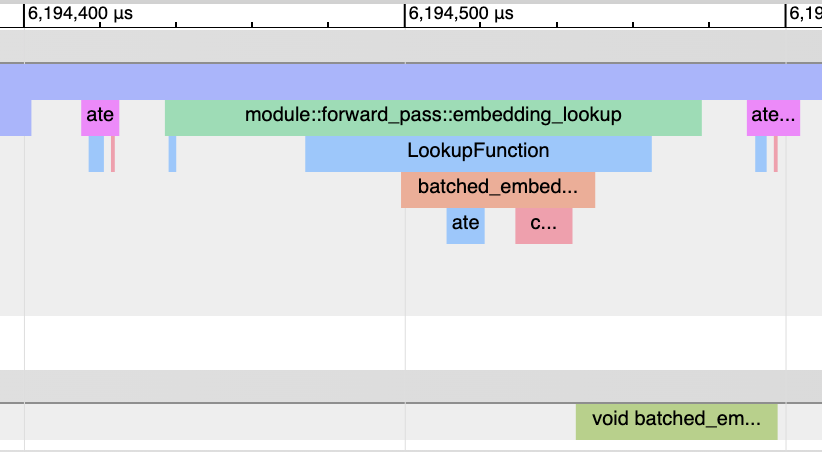}}
  \vskip -0.1in
  \caption{Separate embedding bag ops (left) and batched embedding op (right).}
  \label{fig:emb}
\end{figure}

\paragraph{Op Fusion} Op fusion is a common optimization technique that brings speedup by replacing multiple ops with a mathematically equivalent one to reduce both the compute time and overheads. When users implement a new op, it is good to know how it improves the performance in a ML model generally (i.e., with arbitrary input tensor shapes). Figure~\ref{fig:emb} shows an example of optimization that we have done with the performance model. On the left side it shows a series of embedding bag ops as a good target (i.e., causing too much device overheads) to be fused into a batched embedding op, as shown in the right side.  Our prediction pipeline captures optimization opportunities like this during trace analysis. We then can easily modify the execution graph and replace the subgraph of all embedding bag ops with arbitrary input shapes with one single batched embedding op, whose performance is then predicted by our kernel performance model. This is extremely efficient when there are a large number of ML models to be optimized and evaluated, since we never need to launch jobs and benchmark them.

\paragraph{Load Balancing} In the cases of multi-GPU training, subgraphs that are too expensive to be computed on one single device are distributed to several through data- or model-parallelism. This is also a common practice for DLRM, especially the enormous embedding tables. Our performance model enables the evaluation of each device's performance upon any schemes of splitting embedding tables that results in different combinations of embedding table sizes on these devices. Again, this greatly accelerates the development and debugging of DLRM training on multi-GPU platforms.

\subsection{Extendibility}
Our performance model is designed to be highly extendible for both workloads and devices. To extend, users only need to design/train new kernel performance models and collect op overheads information for the new devices, which is a straightforward and relatively simple effort. To run on different devices, one should also make sure PyTorch's Kineto tracing is able to capture events of kernels running on these new devices in order to support dominating-kernels identification and overheads extraction. Besides, the extension of this work to (distributed) multi-GPU platforms also requires kernel performance models of communication collectives (e.g., \emph{all\_to\_all}, \emph{all\_reduce}). This is one of our work in progress.

\section{Conclusion and Future Work}
We devise a performance model for GPU training of DLRM as well as other ML models.  We find that some ML workloads, with DLRM as a typical example, consist of a broad range of ops and have GPU utilization.Therefore, we propose to use different approaches for constructing kernel performance model for these ops; compared to simply predicting the E2E time as the sum of kernel time,  our work is a more general methodology that covers the case of model configurations with low GPU utilization.  Our final end-to-end performance model is proved to have low error and high extendibility, and is able to assist model-system co-design. Future work includes investigating communication collective performance for modeling ML workload training on (distributed) multi-GPU platforms. Another of our goals is to model the performance of embedding lookups with a non-constant number of embeddings and number of lookups per table, which should improve our overall model accuracy. Finally, we would also love to develop a tool that visualizes and facilitates the manipulation of execution graphs for model-system co-design.

\bibliographystyle{IEEEtran}
\bibliography{performance_model_paper}

\end{document}